\documentclass[letterpaper]{article} 
\usepackage{aaai2027}  

\usepackage[hyphens]{url}  
\usepackage{graphicx} 
\usepackage{natbib}  
\usepackage{caption} 
\usepackage{amsmath,amssymb}

\usepackage{booktabs}
\usepackage{multirow}
\usepackage{makecell}
\usepackage{enumitem}
\usepackage[table]{xcolor}

\title{\resizebox{\textwidth}{!}{Bridging Object Detection and Segmentation with Polygon Detection Transformers}}

\author{
    Jiacheng Sun\textsuperscript{\rm 1,\rm 2},
    Jiaqi Lin\textsuperscript{\rm 1},
    Wenlong Hu\textsuperscript{\rm 1},
    Haoyang Li\textsuperscript{\rm 1},\\
    Xinghong Zhou\textsuperscript{\rm 1},
    Chenghai Mao\textsuperscript{\rm 1},
    Xinliang Zhang\textsuperscript{\rm 2},
    Jianya Guo\textsuperscript{\rm 2},\\
    Yuqiang Zhai\textsuperscript{\rm 2},
    Yan Peng\textsuperscript{\rm 1},
    Xiaomao Li\textsuperscript{\rm 1}\corresponding
}

\affiliations{
    \textsuperscript{\rm 1}Shanghai University
    \qquad
    \textsuperscript{\rm 2}SenseTime Research\\
    lixiaomao@shu.edu.cn
}

\begin{document}

\nocopyright

\maketitle

\begin{abstract}
Box detection and mask segmentation are two dominant paradigms for foreground representation: boxes are efficient but too coarse for object shapes, while masks are accurate but over-modeled for compact geometry. To bridge this gap, we present a Polygon Detection Transformer (Poly-DETR) built upon Polar Representation, where object queries regress a starting point and its fixed number of radial distances to directly construct the contour-approximating polygon. This formulation can be integrated into most DETR-like detectors by linear extension, since box is a degenerate case of Polar Representation with four rays. Furthermore, we propose two simple but necessary designs, Polar Deformable Attention and Position-Aware Training Scheme, to align feature sampling and polygon supervision. As a DETR-oriented advancement of Polar Representation, Poly-DETR outperforms existing polar-based methods by 4.7 mAP on MS COCO. Moreover, we explore the application regimes of polygon detection in geometry-driven domains, including remote sensing, medical imaging, and autonomous driving. In particular, Poly-DETR shows stronger scalability than its mask-based counterpart in high-resolution scenarios. Additional experiments show that, owing to its Transformer structure, Poly-DETR can be naturally extended to recent DETR variants equipped with foundation-model priors.
Codes can be found at \url{https://github.com/Sun15194/Poly-DETR}.

\end{abstract}

\section{Introduction}
\label{sec1:intro}

In the era of foundation models, foreground representation remains a long-standing challenge in visual recognition. As two dominant paradigms, box detection and mask segmentation reflect different representational emphases for foreground objects. Box detection localizes foreground objects with bounding boxes, which is efficient but too coarse to capture boundary cues. In contrast, mask segmentation assigns semantics pixel by pixel, which is more fine-grained but tends to over-model compact geometry, especially in high-resolution scenarios. This motivates a simple yet expressive intermediate representation between boxes and masks.

\begin{figure}[t]
  \centering
  \includegraphics[width=\linewidth]{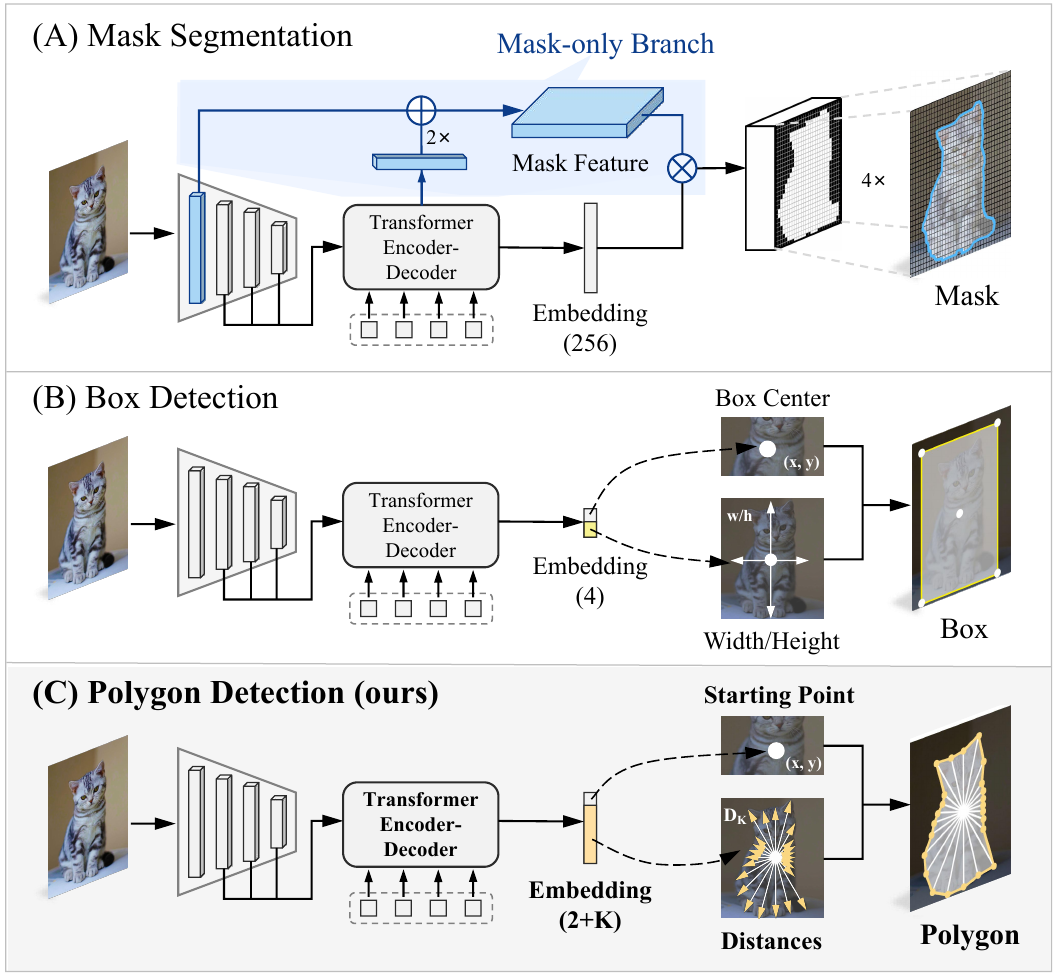}
  \caption{Comparison of three foreground representations in Transformers. Polygon detection offers stronger shape expressiveness by only linearly extending the box-detection network structure, while avoiding the additional pixel-wise branch required by mask segmentation.}
  \label{fig1}
\end{figure}

Recently, the reconsideration of Polar Representation \cite{c15:polarnext} has offered a promising answer to this question. Built on this concept, a bounding polygon that approximates the instance contour can be directly constructed using a fixed number of rays emitted from an interior starting point (as shown in Fig.~\ref{fig1}). In this paper, we introduce Polar Representation into Transformer architectures and propose a Polygon Detection Transformer, termed Poly-DETR. Since bounding box is a degenerate case of Polar Representation with four rays, our design can be integrated into most modern DETR-like detectors \cite{c16:deformable-detr} by linear extension. Moreover, benefiting from the global contextual interaction in Transformers, object queries can directly regress starting points in continuous space, which is more flexible than the classification-driven selection on fixed feature grids used in existing polar-based methods.

\begin{figure}[t]
  \centering
  \includegraphics[width=\linewidth]{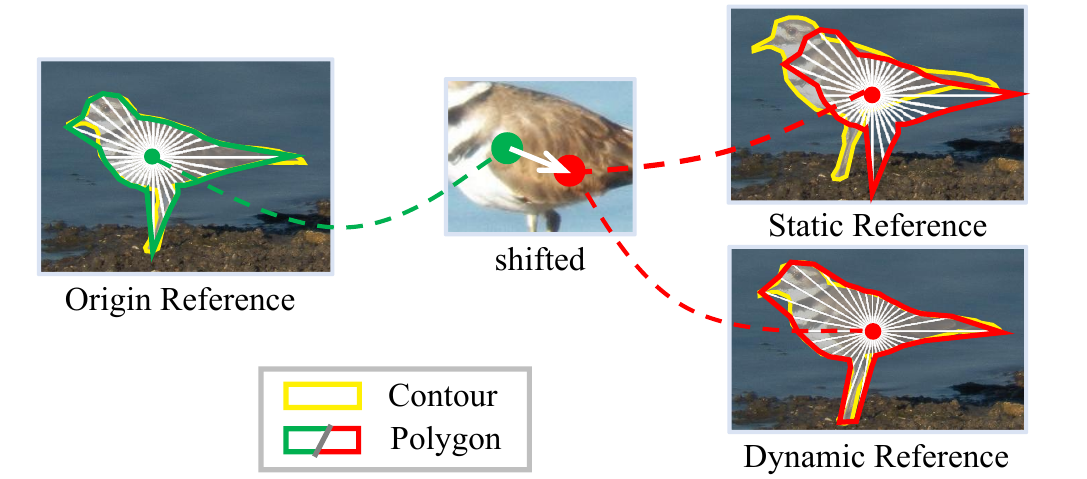}
  \caption{Mismatched Supervision Reference. The starting point shifts from its origin (\textit{green}) to the optimized one (\textit{red}). The static reference yields a polygon that drifts away from the contour (\textit{yellow}), while the dynamic reference keeps the reconstructed polygon aligned.}
  \label{fig2}
\end{figure}

\begin{figure}[t]
  \centering
  \includegraphics[width=\linewidth]{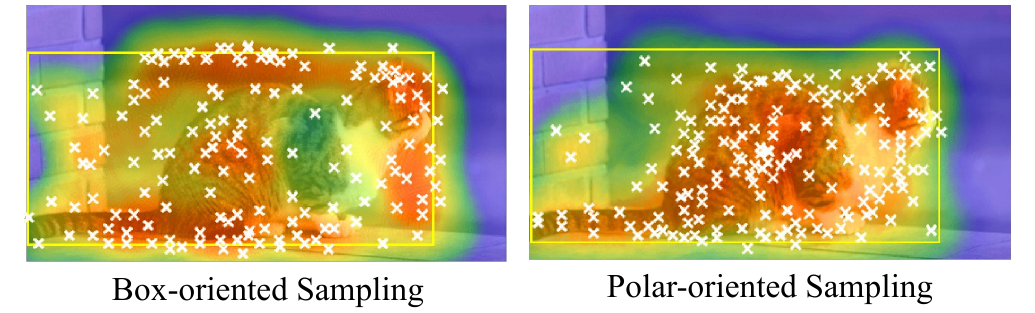}
  \caption{Incompatible Sampling Reference. Deformable Attention sampling locations (\textit{white crosses}) and their density heatmaps show that box-oriented sampling concentrates around box cues, whereas polar-oriented sampling favors boundary regions.}
  \label{fig3}
\end{figure}

However, despite the structural compatibility, polygon detection is not geometrically equivalent to conventional box detection. On the one hand, mainstream DETR-like detectors adopt the box-center encoding, where matched queries always share a fixed reference throughout training, i.e., the center and size of their ground-truth box. In contrast, radial distances are conditioned on starting points, making static polygon supervision misaligned once the predicted points shift (Fig.~\ref{fig2}). On the other hand, Deformable Attention is designed for box-oriented feature extraction, whose sampling locations are anchored by box references. This is suboptimal for polar regression, where informative cues lie around starting points and along instance boundaries (Fig.~\ref{fig3}). To address these issues, we propose two polygon-specific components: a Position-Aware Training Scheme (PATS) that dynamically updates the supervision reference and penalizes outside-contour starting points during matching, and a Polar Deformable Attention (Polar-DA) that constructs fan-shaped sampling grids with learnable offsets scaled by radial distances. Experimental results on MS COCO show that, even using the standard Deformable DETR as baseline, Poly-DETR outperforms state-of-the-art polar-based methods by 4.7 mAP, demonstrating the effectiveness of our approaches.

Another key aspect of this work is to explore the application boundary between polygon detection and mask segmentation, since they delineate the instance shape at different granularities. To this end, we construct a parallel counterpart called Mask-DETR to isolate the effect of representation, which is built under identical data augmentation, network architecture, training schedule and optimizer to Poly-DETR. This controlled comparison leads to two observations. First, as the test-time input resolution is progressively increased on MS COCO, Poly-DETR exhibits flatter scaling curves in computational cost and latency than Mask-DETR. This trend is further validated on the high-resolution Cityscapes dataset (six times that of COCO), where Poly-DETR reduces GPU memory consumption by almost half. Second, on domain-specific datasets with regular-shaped instances, including PanNuke (cell nuclei) and SpaceNet (building footprints), Poly-DETR even surpasses its mask-based counterpart across all evaluation metrics. Beyond these experiments, we further showcase a structured junction topology task in autonomous driving, where explicit supervision for polar angles enables Poly-DETR to directly produce end-to-end polygonal results with semantic endpoints. Moreover, an additional transfer experiment suggests that Poly-DETR can be extended to recent detectors using foundation-model priors, owing to its Transformer architecture. In this setting, our polar-based variant achieves comparable performance to representative mask-based methods, using only a very short training schedule. 

\section{Related Work}
\label{sec2:related}

\subsection{Box Detection}
\label{sec2-1:box}

Box detection localizes foreground objects with rectangular bounding boxes. Classical detectors \cite{c17:faster-rcnn} predict geometric offsets, i.e., box center and size, for predefined anchors. Based on this box-center notion, Detection Transformer \cite{c14:detr} introduces sparse object queries and bipartite matching to remove hand-crafted components. For efficient multi-scale feature interaction, Deformable DETR \cite{c16:deformable-detr} presents a Deformable Attention mechanism to sample and aggregate key features around reference points using learnable offsets. Building on this framework, subsequent methods further explore faster training convergence \cite{c40:mrdetr} and stronger foundation-model priors \cite{c39:rfdetr}.

\begin{figure*}[t]
  \centering
  \includegraphics[width=\linewidth]{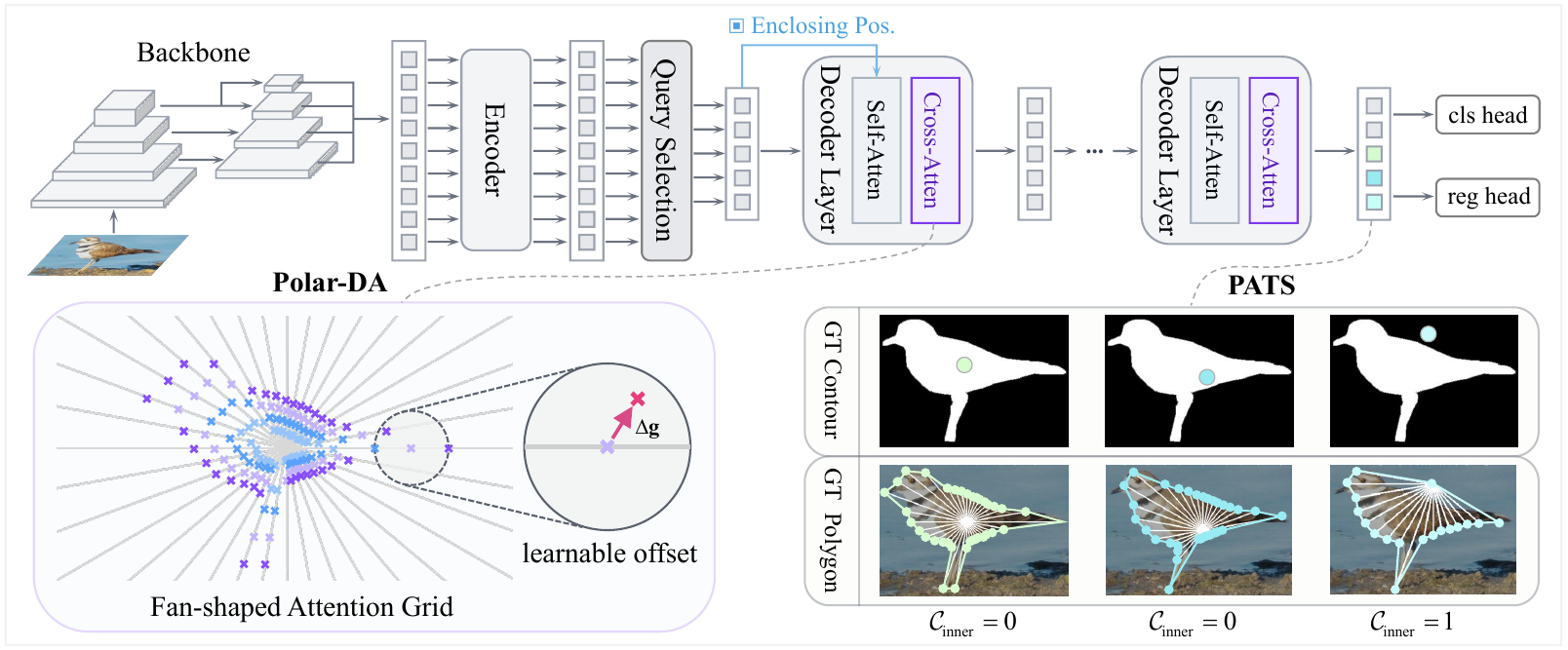}
  \caption{Overview of the proposed Poly-DETR. The upper part illustrates our polygon detection pipeline, where Enclosing Pos. denotes positional embeddings constructed from the enclosing boxes. The bottom-left part shows the sampling mechanism of Polar Deformable Attention. The bottom-right part illustrates the Position-Aware Training Scheme using three color-coded object queries with different predicted starting points.}
  \label{fig4}
\end{figure*}

\subsection{Mask Segmentation}
\label{sec2-2:mask}

Mask segmentation represents instances as dense binary masks. Two-stage methods \cite{c1:maskrcnn} use detected bounding boxes as regions of interest and then perform pixel-wise classification within them. In contrast, one-stage methods \cite{c2:yolact, c4:sparseinst} construct a shared mask feature map over the entire image and predict instance-specific kernels for mask generation via dot product or convolution. Transformer-based methods \cite{c5:mask2former, c6:maskdino} further extend this one-stage paradigm by using object queries to parameterize mask embeddings, together with additional mask branches to produce mask features.

\subsection{Polar Representation}
\label{sec2-3:polar}

Polar Representation conceptually enables polygon detection using a starting point and its ordered radial distances. This paradigm is first explored by ESE-Seg \cite{c38:eseseg} and later integrated into classical box detectors by PolarMask \cite{c12:polarmask, c13:polarmask++}. However, Polar Representation remains underexplored due to its performance gap with mask-based methods. Recently, PolarNeXt \cite{c15:polarnext} revisits this paradigm by revealing an inherent representation error and proposes RMask IoU for assessment. In this paper, we explicitly position Polar Representation as an intermediate representation between boxes and masks, and further investigate how it scales to modern Transformer architectures while systematically characterizing the application regimes. Unlike previous methods, Poly-DETR models starting points in continuous space rather than restricting them to discrete feature-map grids, which is particularly effective in avoiding suboptimal starting-point placement.

\subsection{Contour Refinement}
\label{sec2-4:contour}

Contour refinement is a line of work that resembles polygon detection at the output level, as both represent instances with vertex sets, but differs fundamentally in formulation. Following the Active Contour Model \cite{c23:deepsnake}, contour-based methods \cite{c24:e2ec, c26:dance, c25:polysnake} are built upon box detectors with additional multi-stage networks, which progressively deform initial shapes toward instance contours. Recent methods \cite{c27:boundaryformer, c41:contourformer} further introduce Transformer architectures for boundary modeling. In contrast, polygon detection approximates instance contours in a single-shot manner, while the ordered rays in polar coordinates naturally alleviate topological issues in Cartesian vertex regression, such as self-intersection.

\section{Method}
\label{sec3:method}

\subsection{Polar Parameterization}
\label{sec3-1:parms} 

For a given instance contour $\mathbf{C}$, Polar Representation $\psi$ constructs its approximating polygon $\mathbf{P}=\psi(\mathbf{s},\mathbf{D})$ from a starting point $\mathbf{s}=[x,y]^\top\in\mathbb{R}^2$ and $K$ radial distances $\mathbf{D}=[d_1,\ldots,d_K]^\top\in\mathbb{R}_+^K$ measured along uniformly spaced rays. In DETR-like detectors, sparse object queries directly regress 4-dimensional box parameters $\mathbf{b}=[x,y,w,h]^\top$, where $(x,y)$ denotes the box center and $(w,h)$ denotes the width and height. Following the same query-based paradigm, Poly-DETR regresses $(2+K)$-dimensional polar parameters in the form of:
\begin{equation}
    \label{eq1:polar-param}
    \mathbf{p} = [\mathbf{s}; \mathbf{D}]
    = [x,\; y,\; d_1, \ldots, d_K]^{\top}
    \in \mathbb{R}^{2 + K}.
\end{equation}
In this setting, the migration of our polar-based formulation to Transformer architectures incurs linear overhead, mainly arising from the expansion of the regression head. 

However, despite the structural compatibility with Detection Transformers, polygon detection is not geometrically equivalent to box detection. Modern DETR-like detectors are typically built upon the box-center notion, since the box center and size are stable geometric references throughout training. In contrast, radial distances in Polar Representation
are conditioned on the current predicted starting point. Consequently, the box-oriented static supervision and feature sampling result in two geometric mismatches illustrated in Fig.~\ref{fig2} and Fig.~\ref{fig3}, which motivate two simple but necessary designs introduced in the following two sections.

\subsection{Position-Aware Training Scheme}
\label{sec3-2:pats}

To make the integration of Polar Representation clear and reproducible on Detection Transformers, we instantiate Poly-DETR on standard Deformable DETR, which serves as a pure and representative baseline for subsequent DETR-like models. As illustrated in Fig.~\ref{fig4}, Poly-DETR retains the standard pipeline, consisting of a backbone, a Transformer encoder, a Transformer decoder, and a group of MLP prediction heads. However, in Polar Representation, radial distances are conditioned on the current starting point, while optimization during training tends to shift the predicted starting point toward locations with lower representation errors. This invalidates the conventional formulation that defines regression targets as static geometric attributes anchored to a fixed reference point, such as the box center or instance centroid. Therefore, we propose a \textbf{Position-Aware Training Scheme} (PATS) to dynamically update the supervision reference of radial distances according to the predicted starting point. For a prediction matched to an instance contour $\mathbf{C}$, radial distances $\mathbf{D}^{\mathrm{ref}}$ used for reference can be formulated as:
\begin{equation}
    \label{eq2:pats-distance}
    d_k^{\mathrm{ref}}
    =
    \max
    \left\{
    \rho \geq 0 \mid
    \operatorname{sg}(\hat{\mathbf{s}}) + \rho \mathbf{u}_k
    \in \partial \mathcal{A}(\mathbf{C})
    \right\},
\end{equation}
\begin{equation}
    \label{eq3:pats-vector}
    \mathbf{D}^{\mathrm{ref}}
    =
    [d_1^{\mathrm{ref}},\ldots,d_K^{\mathrm{ref}}]^\top,
\end{equation}
where $\hat{\mathbf{s}}$ denotes the predicted starting point, $\mathbf{u}_k=[\cos\theta_k,\sin\theta_k]^\top$ is the unit vector of the $k$-th ray, $\mathcal{A}(\mathbf{C})$ denotes the instance region enclosed by $\mathbf{C}$, and $\operatorname{sg}(\cdot)$ stops the gradient through the reference construction. Built on this design, Distance L1 Loss supervises the position-conditioned radial distances for internal regression constraint, while RMask IoU Loss jointly optimizes
starting point coordinates and radial distances for external shape constraint. 

Furthermore, once supervision is dynamically updated, two native components of DETR detectors must be adapted accordingly to ensure the effectiveness of PATS:

\noindent\textbf{(1)} Unlike CNN-based methods that rely on dense center priors for sample assignment, DETR-like detectors perform sparse query matching without explicit inside-contour constraint on starting points. At the early training stage, a substantial proportion of outside points results in large representation errors and thus unreliable positive/negative assignments. To address this issue, we introduce an Inner Cost into bipartite matching to encourage more queries with interior points to be selected as positives:
\begin{equation}
    \label{eq4:inner-cost}
    \mathcal{C}_{\mathrm{inner}}
    =
    \begin{cases}
    1, & \hat{\mathbf{s}} \notin \mathcal{A}(\mathbf{C}) \\
    0, & \hat{\mathbf{s}} \in \mathcal{A}(\mathbf{C})
    \end{cases}.
\end{equation}

\noindent\textbf{(2)} The position-aware reference complicates the encoder-side construction of positional embeddings for object queries. Critically, without instance-level interaction, radial-distance predictions from encoder features cannot provide reliable instance geometry. To this end, we replace the full polar reference with its enclosing rectangle, which provides more stable extent cues while retaining the starting-point coordinates:
\begin{equation}
    \label{eq5:xyltrb-pos}
    \begin{gathered}
    \mathbf{q}_{\mathrm{pos}}
    =
    \operatorname{FC_{256}}
    \left(
    \operatorname{PE}_{128}(x,y);
    \operatorname{PE}_{64}(l,t,r,b)
    \right),
    \end{gathered}
\end{equation}
where $(x,y)$ denote starting points, and $(l,t,r,b)$ are their distances to the four edges of rectangles. $\operatorname{PE}_{128/64}(\cdot)$ denotes positional encoding with 128/64 channels for each scalar input, and $\operatorname{FC}_{256}(\cdot)$ is a fully connected layer that projects the concatenated positional embedding to 256 channels.

\subsection{Polar Deformable Attention}
\label{sec3-3:polar-da}

Besides supervision mismatch, another challenge for Polygon Detection Transformers lies in the incompatible sampling reference of box-oriented attention mechanisms. In the decoder structure of modern DETR-like detectors, cross-attention commonly adopts Deformable Attention (termed Vanilla-DA), which sparsely samples and aggregates multi-scale features. For each object query ${{\bf{q}}^{[l]}}$ at decoder layer $l$, given a 4D box reference $(x,y,w,h)$, Vanilla-DA predicts a small set of sampling offsets $\Delta {\bf{g}} \in {\mathbb{R}^2}$ for each attention head. Concretely, the sampling location is obtained by adding offsets $\Delta {\bf{g}}$ to the reference center $(x,y)$, while $\Delta {\bf{g}}$ is rescaled by the box size $(w,h)$, i.e., ${\bf{g}} = (x,y) + \Delta {\bf{g}} \odot (w,h)$, and features are then bilinearly sampled at ${\bf{g}}$. This box-oriented sampling typically converges to a box-shaped pattern, where samples cluster around the box center and become denser near box edges (as shown in Fig.~\ref{fig3}), which aligns well with box regression and classification. 

However, for polygon detection based on Polar Representation, sampling locations should instead focus on the neighborhood of the starting point and boundary evidence along each radial direction. To geometrically align the cross-attention aggregated features with the regression targets of Polar Representation, we propose \textbf{Polar Deformable Attention} (Polar-DA) to re-parameterize the sampling scheme in a polar-oriented manner. To be specific, cross-attention reference is shifted from the box center $\bf{c}$ to starting point $\bf{s}$, so that the sampling anchor matches the geometric origin of radial-distance regression, while reducing the systematic bias of box-centered sampling toward interior cues. In terms of head semantics, each polar ray direction is paired with one attention head, i.e., $K$ heads for $K$ rays, thereby endowing attention heads with explicit direction-aware semantics. Intuitively, sampling four points per direction yields a fan-shaped grid as shown in the bottom-left of Fig.~\ref{fig4}. This grid can also be formalized as base sampling locations $\overline {\bf{g}}$ along each ray direction:
\begin{equation}
\label{eq6:poly-grid}
    \overline{\mathbf{g}}_{k,t}^{[l]}
    =
    \mathbf{s}^{[l-1]}
    +
    \frac{t}{T} d_k^{[l-1]} \mathbf{u}_k ,
\end{equation}
where $k$ and $t$ index ray directions and sampling points, respectively, ${{\bf{u}}_k} \in {\mathbb{R}^2}$ is the unit direction vector and $T$ is the number of sample points. The final sampling points ${\bf{g}}$ are obtained by adding learnable offsets $\Delta {\bf{g}}$ on top of the fan-shaped grid, with the offsets modulated by $d_k^{[l-1]}$: 
\begin{equation}
\label{eq6:poly-sampling}
    {\bf{g}}_{k,t}^{[l]} = \overline {\bf{g}} _{k,t}^{[l]} + \Delta {\bf{g}}_{k,t}^{[l]} \odot \frac{{d_k^{[l-1]}{{\bf{u}}_k}}}{T}.
\end{equation}
Compared with the box-scale modulation $(w,h)$ in vanilla Deformable Attention, our distance-conditioned scaling by $d_k^{[l-1]}$ stabilizes sampling around the starting-point neighborhood and near the instance boundary.

\section{Experiments}
\label{sec4:experiment}

\subsection{Experimental Settings}
\label{sec4-1:exp-set}

In this section, main results and ablation studies are conducted on MS COCO \cite{c18:mscoco}, setting the shorter side to 800 pixels with aspect ratio preserved. Following common practice for fair comparison, data augmentation only contains random flip and scale jitter unless specified. In addition, Cityscapes \cite{c19:cityscapes} is introduced for high-resolution evaluation, while PanNuke \cite{c9:pannuke} and SpaceNet \cite{c10:spacenet} are used to study modeling of regular-shaped instance contours. In terms of evaluation metrics, the segmentation accuracy is evaluated by the standard Mask mAP, with $\rm{AP_{50}}$ and $\rm{AP_{75}}$ also reported. All experiments are conducted on NVIDIA RTX 4090, using 4 GPUs for training and 1 GPU for inference. Moreover, inference speed, model complexity, computational overhead and memory consumption are respectively measured by FPS, Params (M), FLOPs (G) and Mems (MB). TensorRT or FP16 is not used for acceleration.

\begin{table}[t]
\centering
\setlength{\tabcolsep}{3pt}
\begin{tabular}{lccc}
\toprule
\textbf{Comment} & $\mathbf{mAP}$ & $\mathrm{\mathbf{AP}}_\mathbf{50}$ & $\mathrm{\mathbf{AP}}_\mathbf{75}$ \\
\midrule
Baseline   & 33.4 & 56.5 & 34.3 \\
+DETR      & 34.3 (+0.9) & 60.6 & 34.2 \\
+Polar-DA  & 36.1 (+2.7) & 61.5 & 36.4 \\
\rowcolor{gray!10}
+PATS      & 37.8 (+4.4) & 62.2 & 39.1 \\
\bottomrule
\end{tabular}
\caption{Ablation study on module components and their impact on accuracy.}
\label{tab1:ablation_component}
\end{table}

\begin{table}[t]
\centering
\setlength{\tabcolsep}{4pt}
\begin{tabular}{cccc}
\toprule
\textbf{Type} & \textbf{H/T/R} & \textbf{mAP} & \textbf{Params} \\
\midrule
\multirow{2}{*}{Vanilla-DA}
& 8/4/1.0   & 36.1        & 42.39 \\
& 32/4/1.0  & 36.2 (+0.1) & 43.68 \\
\midrule
& 32/4/4.0  & 38.1 (+2.0) & 50.21 \\
\rowcolor{gray!10}
& 32/4/1.0
& 37.8 (+1.7)
& 43.68 \\
\multirow{-3}{*}{Polar-DA}
& 32/2/1.0
& 37.5 (+1.4)
& 43.17 \\
\bottomrule
\end{tabular}
\caption{Ablation study on attention design. \textbf{H}: attention heads; \textbf{T}: sampling points; \textbf{R}: projection ratio.}
\label{tab2:ablation_attention}
\end{table}

\subsection{Implementation Details}
\label{sec4-2:imp-detail}

For the experiments in this section, Poly-DETR is built upon a Deformable DETR baseline with the same model configuration, using a ResNet-50 backbone pretrained on ImageNet and a 6-layer encoder-decoder Transformer. All models are implemented on MMDetection toolbox, optimized by AdamW with a base learning rate of 1e-4. In the decoder, object queries are trained with Classification Focal Loss, Distance L1 Loss, and RMask IoU Loss. Hungarian matching is performed using the corresponding classification, distance, and RMask costs, together with an inner cost. The loss weights $\{\lambda_{\mathrm{cls}}, \lambda_{\mathrm{dist}}, \lambda_{\mathrm{rmask}}\}$ are set to $(2, 5, 2)$, and the inner-cost weight is set to $\lambda_{\mathrm{inner}}=5$. The number of rays in Polar Representation is set to $K = 32$ (power of 2) to align with channel allocation. Following the hybrid supervision strategy \cite{c29:msdetr}, we use 12 and 36 epochs as the short and long training schedules, respectively. More details are provided in supplementary materials.

\subsection{Ablation Studies}
\label{sec4-3:ablation}

For a fair comparison, all ablation experiments are conducted under the 12-epoch training schedule on MS COCO val.

\noindent \textbf{Component Impact.} The incremental impact of the proposed components is reported in Tab.~\ref{tab1:ablation_component}. The straightforward migration to Transformer architectures (+DETR) improves mAP by 0.9 over the PolarNeXt baseline, indicating that stronger classification and localization directly benefit polygon detection. With Polar Deformable Attention (+Polar-DA), the gain further expands to 2.7 mAP, validating the effectiveness of the fan-shaped sampling pattern for distance regression. By incorporating Position-Aware Training Scheme (+PATS), the overall performance reaches 37.8 mAP (+4.4), with $\rm{AP_{75}}$ improving notably from 34.3 to 39.1. These results underscore that, while the straightforward Transformer migration brings certain gains, achieving the optimal performance still relies on the proposed polygon-specific designs.

\noindent \textbf{Attention Design.} Tab.~\ref{tab2:ablation_attention} investigates how the number of attention heads \textbf{H} and sampling points per head \textbf{T}, as well as the channel projection ratio \textbf{R} (the ratio of per-head feature dimension relative to Vanilla-DA), affect performance and model complexity. Rows~1-2 show that simply using more attention heads in Vanilla-DA brings little gain while increasing complexity. Row~3 confirms that the full-capacity Polar-DA configuration improves mAP by 2.0 over Vanilla-DA, but incurs substantial parameter overhead due to the increased per-head feature dimension. This overhead can be alleviated by reducing the projection ratio to match the Vanilla-DA width, resulting in only a slight mAP drop in Row~4. Conversely, reducing the sampling density in Row~5 causes another 0.3 mAP drop while saving only a small number of parameters. Overall, the 32/4/1.0 configuration in Row~4 is adopted as a better trade-off in this paper.

\noindent \textbf{Position-Aware Design.} Tab.~\ref{tab3:pats_ablation} decomposes PATS into three parts and studies their complementarity. Experimental results show that dynamic supervision alone improves mAP by 0.4, suggesting that updating radial-distance targets is necessary but insufficient. When the positional embedding and matching cost are further optimized, these designs bring progressive gains and yield a cumulative improvement of 1.7 mAP. Moreover, the last row reveals a clear coupling among them: the positional embedding and matching cost are beneficial only when supervision is dynamically aligned. Otherwise, these position-aware cues become inconsistent with instance contours and even degrade performance below the baseline.

\begin{table}[t]
\centering
\setlength{\tabcolsep}{4pt}
\begin{tabular}{cccc}
\toprule
\textbf{Dynamic} & \textbf{Enclosing} & \textbf{Inner} & \textbf{mAP} \\
\midrule
 &  &  & 36.1 \\
\checkmark &  &  & 36.5 ($+$0.4) \\
\checkmark & \checkmark &  & 37.3 ($+$1.2) \\
\rowcolor{gray!10}
\checkmark & \checkmark & \checkmark & 37.8 ($+$1.7) \\
 & \checkmark & \checkmark & 35.2 ($-$0.9) \\
\bottomrule
\end{tabular}
\caption{Ablation study on Position-Aware Training Scheme. 
\textit{\textbf{Dynamic}}, \textit{\textbf{Enclosing}}, and \textit{\textbf{Inner}} denote dynamic supervision, enclosing rectangle for positional embedding, and inner cost, respectively.}
\label{tab3:pats_ablation}
\end{table}

\begin{table*}[t]
\centering
\setlength{\tabcolsep}{5pt}
\renewcommand{\arraystretch}{1.05}
\begin{tabular}{clcccccccc}
\toprule
\textbf{Type} & \textbf{Method} & \textbf{Backbone} & \textbf{Epoch} 
& \textbf{mAP} & \textbf{AP50} & \textbf{AP75} 
& \textbf{FPS} & \textbf{Params$\downarrow$} & \textbf{FLOPs$\downarrow$} \\
\midrule
\multirow{6}{*}{\shortstack{Contour}}
& DeepSnake       & DLA34   & 160 & 30.3 & --   & --   & 24 & --    & --  \\
& E2EC            & DLA34   & 150 & 33.8 & 52.9 & 32.8 & 35 & 29.80 & 121 \\
& PolySnake       & DLA34   & 250 & 34.4 & --   & --   & 18 & --    & --  \\
& DANCE           & Res50   & 36  & 36.8 & 58.5 & 39.0 & 16 & 44.60 & 274 \\
& BoundaryFormer  & Res50   & 12  & 36.4 & 57.2 & 39.0 & 19 & 46.32 & 229 \\
& ContourFormer   & HGNetv2 & 72  & 37.6 & 58.1 & 39.7 & 21 & --    & --  \\
\midrule
\multirow{5}{*}{\shortstack{Polygon}}
& ESE-Seg         & Res50   & 300 & 21.6 & 48.7 & 22.4 & 38 & --    & --  \\
& PolarMask       & Res50   & 36  & 31.3 & 52.5 & 32.3 & 42 & 34.74 & 264 \\
& PolarNeXt       & Res50   & 36  & 36.1 & 59.7 & 37.3 & 49 & 32.36 & 186 \\
\rowcolor{gray!10}
\cellcolor{white} & \textbf{Poly-DETR (ours)} & Res50 & 12 
& 38.1 & 62.5 & 39.5 & 32 & 43.68 & 208 \\
\rowcolor{gray!10}
\cellcolor{white} & \textbf{Poly-DETR (ours)} & Res50 & 36 
& 40.8 & 66.9 & 42.7 & 32 & 43.68 & 208 \\
\bottomrule
\end{tabular}
\caption{Comparison with contour/polar-based methods on MS COCO test-dev. 
``$\downarrow$'' means smaller values are preferred.}
\label{tab4:compare2poly}
\end{table*}

\begin{table}[t]
\centering
\small
\setlength{\tabcolsep}{3pt}
\renewcommand{\arraystretch}{1.05}
\begin{tabular}{lcccccc}
\toprule
\textbf{Method} & \textbf{mAP} & $\mathbf{AP}_{50}$ & $\mathbf{AP}_{75}$ 
& \textbf{FPS} & \textbf{FLOPs$\downarrow$} & \textbf{Mem$\downarrow$} \\
\midrule
\multicolumn{7}{c}{\textbf{MS COCO val (800$\times$1333)}} \\
\midrule
Mask-DETR & 39.8 & 62.9 & 43.1 & 24 & 264 & 643 \\
\rowcolor{gray!10}
Poly-DETR & 37.8 & 63.1 & 39.5 & 32 & 208 & 521 \\
\midrule
\multicolumn{7}{c}{\textbf{Cityscapes (1024$\times$2048)}} \\
\midrule
Mask-DETR & 35.3 & 59.7 & 34.1 & 10 & 533 & 1557 \\
\rowcolor{gray!10}
Poly-DETR & 33.8 & 62.4 & 29.7 & 15 & 454 & 833 \\
\midrule
\multicolumn{7}{c}{\textbf{PanNuke (256$\times$256)}} \\
\midrule
Mask-DETR & 20.8 & 49.5 & 13.3 & 39 & 24 & 229 \\
\rowcolor{gray!10}
Poly-DETR & 24.6 & 53.7 & 18.6 & 41 & 22 & 217 \\
\midrule
\multicolumn{7}{c}{\textbf{SpaceNet (650$\times$650)}} \\
\midrule
Mask-DETR & 48.4 & 81.1 & 52.7 & 25 & 127 & 463 \\
\rowcolor{gray!10}
Poly-DETR & 49.0 & 82.2 & 53.9 & 32 & 113 & 340 \\
\bottomrule
\end{tabular}
\caption{Comparison between Poly-DETR and Mask-DETR across datasets. All results are evaluated using 12-epoch models. Poly-DETR and Mask-DETR contain 43.68M and 49.39M parameters, respectively.}
\label{tab5:compare_mask}
\end{table}

\subsection{Comparison with Contour/Polar-based Methods}
\label{sec4-4:comparison2poly}

As shown in Tab.~\ref{tab4:compare2poly}, we compare Poly-DETR with representative methods that explicitly model instances with vertex sets on MS COCO test-dev. Contour refinement methods are included for a comprehensive comparison since they are similar at the output level. Poly-DETR demonstrates a clear advantage over contour-based methods without relying on additional components for progressive deformation. In particular, compared with BoundaryFormer \cite{c27:boundaryformer} and ContourFormer \cite{c41:contourformer} that are also built on Transformers, Poly-DETR achieves higher mAP using only a ResNet-50 backbone and a 12-epoch training schedule. Among polar-based methods, Poly-DETR further improves upon the state-of-the-art PolarNeXt by 4.7 mAP and delivers an even larger gain on the stricter AP$_{75}$ (+5.4), indicating substantially better boundary quality at high IoU thresholds. Although Poly-DETR increases latency and complexity due to the relatively heavy Deformable DETR baseline, this overhead can be largely alleviated when lightweight DETR designs are adopted in Tab.~\ref{tab6:rfdetr_poly}. 

\begin{figure}[t]
  \centering
  \includegraphics[width=0.8\linewidth]{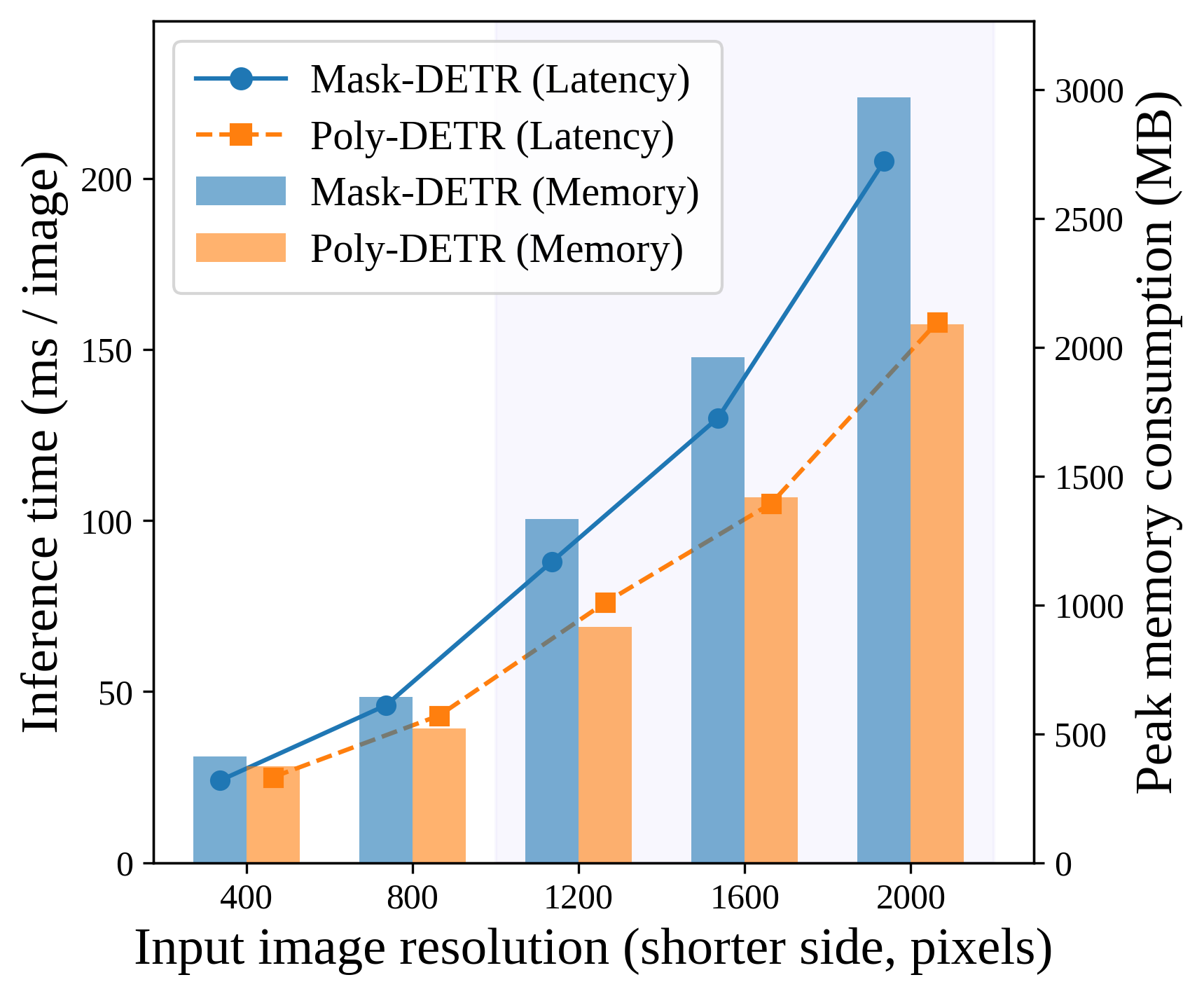}
  \caption{High-resolution scalability on MS COCO. Poly-DETR shows lower inference latency and peak memory consumption than Mask-DETR as the input resolution increases.}
  \label{fig5}
\end{figure}

\subsection{Comparison with Mask-based Methods}
\label{sec4-5:comparison2mask}

We further compare Poly-DETR with a mask-based counterpart to explore the application boundary between polygon detection and mask segmentation. To avoid a misleading “Polygon beats Mask” impression, we construct a parallel Mask-DETR to isolate the effect of representation, rather than competing with advanced mask-based methods by adopting highly engineered baselines. Mask-DETR follows the mask prediction paradigm used in Mask2Former \cite{c5:mask2former} and Mask DINO \cite{c6:maskdino}, while keeping the same data augmentation, network architecture, training schedule and optimizer as Poly-DETR. More details are provided in supplementary materials.

\noindent \textbf{Overall Comparison.}
Tab.~\ref{tab5:compare_mask} summarizes the overall comparison between Poly-DETR and Mask-DETR across datasets. Poly-DETR is consistently more efficient, with higher FPS, lower FLOPs, and reduced peak memory than its mask-based counterpart. On generic datasets such as MS COCO and Cityscapes, Poly-DETR is slightly lower in mAP and AP75, but achieves higher AP50 with clear efficiency advantages. This suggests that dense mask prediction remains stronger for fine-grained boundary modeling on arbitrary objects, while polygon detection provides a more compact representation with better scalability. Based on these observations, we further examine the application regimes of polygon detection in two typical settings: high-resolution inputs and regular-shaped instances.

\noindent \textbf{High-Resolution Scalability.}
In Fig.~\ref{fig5}, we conduct a cross-resolution inference study to examine how the two representations scale with increasing input resolution. Under a constrained accuracy gap ($\Delta$mAP $\leq$ 2), Poly-DETR exhibits smoother growth curves in both latency and peak memory than Mask-DETR. This trend is expected to become more pronounced with lightweight DETR baselines, where the dense mask branch would contribute more noticeably to the overall inference cost. Beyond this controlled resolution-scaling study, we further evaluate the two models on Cityscapes dataset in Tab.~\ref{tab5:compare_mask}, whose native image size is over six times that of the original COCO images. Poly-DETR reduces the peak memory from 1557MB to 833MB ($46.5\%\downarrow$) and improves inference speed from 10 FPS to 15 FPS, confirming its scalability in high-resolution scenarios.

\begin{figure}[t]
  \centering
  \includegraphics[width=\linewidth]{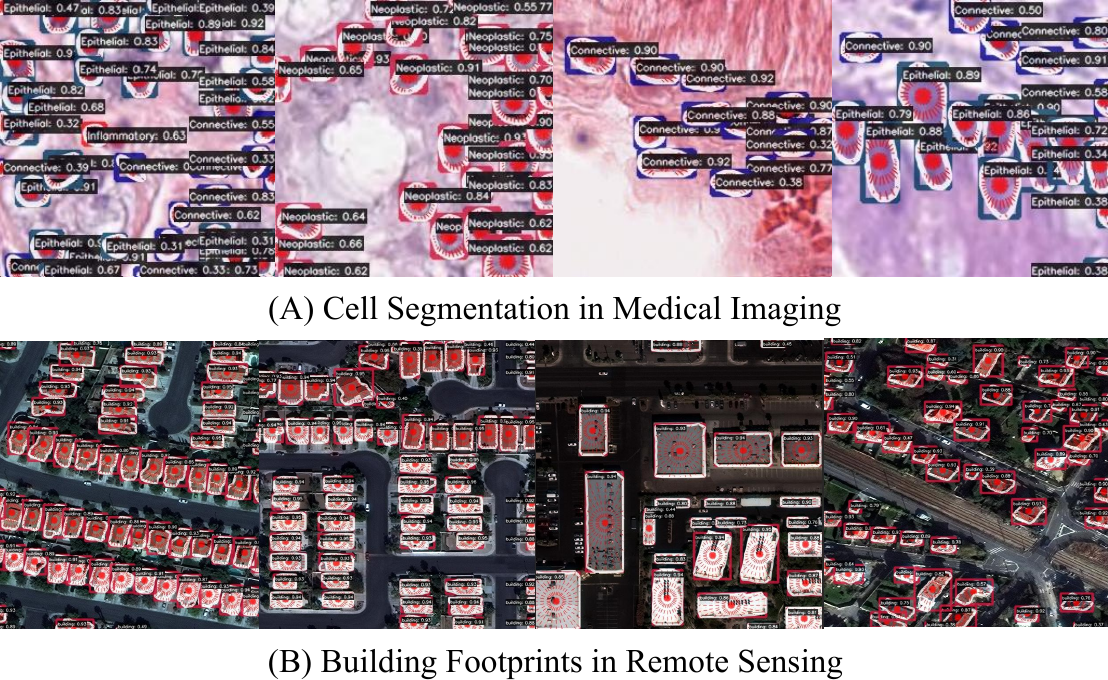}
  \caption{Qualitative results of Poly-DETR on regular-shaped instance domains.}
  \label{fig6}
\end{figure}

\noindent \textbf{Regular-Shaped Instances.}
As shown in Tab.~\ref{tab5:compare_mask}, we extend the comparison to two real-world domains where instances are inherently more regular: medical imaging and remote sensing. On PanNuke for cell segmentation and SpaceNet for building footprints, Poly-DETR consistently outperforms Mask-DETR. It improves mAP from 20.8 to 24.6 on PanNuke and from 48.4 to 49.0 on SpaceNet, while also being faster and more lightweight. These results confirm that polygon detection can serve as a strong alternative to masks for elliptical or orthogonal instances commonly found in such domains. The qualitative results in Fig.~\ref{fig6} further show that Poly-DETR produces compact polygonal contours for both elliptical cell nuclei and orthogonal building footprints.

\section{Polygon as Geometric Interfaces}
\label{sec5:geo_interface}

Beyond standard detection and segmentation settings, this section presents two case studies to show how Polygon Detection Transformers can serve as geometric interfaces.

\begin{figure}[t]
  \centering
  \includegraphics[width=\linewidth]{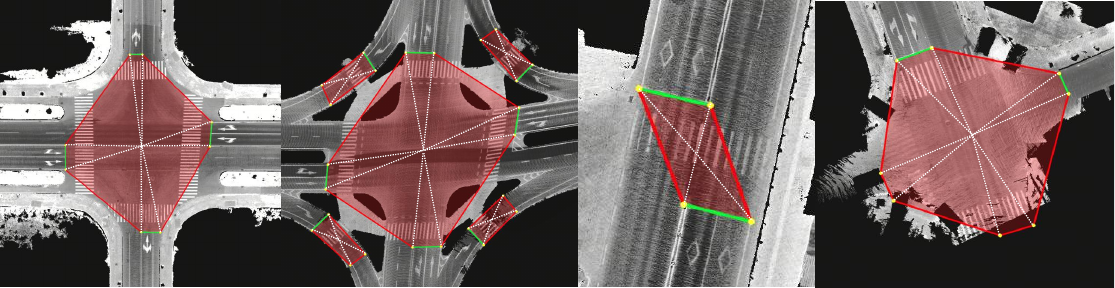}
  \caption{Qualitative results of Poly-DETR for structured junction topology prediction in autonomous driving tasks. Green solid lines denote stop lines, while white dashed lines indicate rays parameterized by learnable polar angles.}
  \label{fig7}
\end{figure}

\begin{table}[t]
\centering
\small
\resizebox{\columnwidth}{!}{
\begin{tabular}{lcccc}
\toprule
Methods          & Epoch & mAP & FPS & Params$\downarrow$ \\
\midrule
RF-DETR-M-seg    & 12 & 44.6 & 37 & 36.20 \\
\rowcolor{gray!10}
RF-DETR-M-poly   & 12 & 43.2 & 42 & 33.90 \\
Mask2Former-R50  & 50 & 42.9 & 23 & 44.03 \\
\bottomrule
\end{tabular}
}
\caption{Comparison of RF-DETR-based and Mask2Former mask/polygon variants on MS COCO val.}
\label{tab6:rfdetr_poly}
\end{table}

\subsection{Geometric Interface for Structured Prediction}
\label{sec5-1:structured_task}

We first introduce a novel downstream case of structured junction topology prediction for autonomous driving. In this application, polygons constructed by Polar Representation serve as geometric carriers that organize the intersection area, crosswalk-adjacent opposing lanes and right-turn regions into a unified topology. Their vertices are defined by the endpoints of related stop lines, which provides explicit semantic supervision for polar parameters during training. This enables the network to predict polar angles as learnable regression parameters together with radial distances. As shown in Fig.~\ref{fig7}, Poly-DETR directly outputs polygons in an end-to-end manner, where the vertices localize related stop lines and the ordered polygonal structure encodes their topological associations. Unlike mask segmentation, this formulation produces structured geometry without extracting contours and endpoints from dense pixel-level predictions through post-processing.

\subsection{Geometric Interface for Foundation-Model Priors}
\label{sec5-2:foundation_prior}

Having migrated Polar Representation to Transformers, we further examine its compatibility with recent DETRs enhanced by foundation-model priors. We adopt RF-DETR-M \cite{c39:rfdetr}, which incorporates internet-scale visual priors through a DINOv2-initialized ViT backbone. For a controlled comparison, we construct two variants equipped with their representation-specific designs: RF-DETR-M-poly and RF-DETR-M-seg. As shown in Tab~\ref{tab6:rfdetr_poly}, our polygon variant further narrows the performance gap with its mask-based counterpart to 1.4 mAP, while achieving lower latency and fewer parameters. We conjecture that this observation arises from the combined engineering designs, such as stronger augmentation, denoising strategies, or large-model distillation, which may transfer more effectively to polygon regression since it retains the original detection-style formulation. More importantly, benefiting from foundation-model priors, RF-DETR-M-poly surpasses the representative mask-based method Mask2Former across all reported metrics (no FP16 or TensorRT), even using a shorter training schedule.

\bibliography{aaai2027}

\appendix
\setcounter{secnumdepth}{1}

\twocolumn[
\begin{center}
    {\LARGE\bfseries Appendix}
\end{center}
\vspace{0.5em}
]

Here we provide more details on the design principles and experimental results for the main paper. An overview of this supplementary material is as follows:
\begin{itemize}[leftmargin=*,labelsep=3mm,label=\textbullet]
    \item Section \ref{app:sec1}: Quick Start for Key Concepts.
    \item Section \ref{app:sec2}: Beyond Transformer Migration.
    \item Section \ref{app:sec3}: More Details for Poly-DETR.
    \item Section \ref{app:sec4}: More Details for Controlled Analysis.
    \item Section \ref{app:sec5}: Training Acceleration with Box Branches.
    \item Section \ref{app:sec6}: Qualitative Visualizations.
    \item Section \ref{app:sec7}: Error Analysis.
\end{itemize}


\section{Quick Start for Key Concepts}
\label{app:sec1}

\subsection{Polar Representation}
\label{app:sec1-1}

PolarMask \cite{c12:polarmask} introduces Polar Representation, which reformulates foreground representation as the prediction of a starting point and a set of radial distances. Given an instance contour $\mathbf{C}$, the approximate polygon $\mathbf{P}$ can be constructed from an arbitrary point $\mathbf{s}$, referred to as \textbf{Starting Point}. Concretely, the point $\mathbf{s}=(x,y)$ is used as the pole of a polar coordinate system, and $K$ rays are emitted at uniformly spaced angles with interval $\theta$ (i.e., $\theta_i=\frac{2\pi(i-1)}{K}$). The intersection point between each ray and $\mathbf{C}$ yields a vertex of $\mathbf{P}$:
\begin{equation}
\label{app:eq1}
    \left\{
    \begin{aligned}
    x_i &= x + d_i \cos \theta_i\\
    y_i &= y + d_i \sin \theta_i
    \end{aligned},
    \right.
\end{equation}
where \textbf{Radial Distances} are defined as $\mathbf{D}=\{d_i\}_{i=1}^{K}$, with $d_i$ measuring the distance from its starting point to the $i$-th intersection. Notably, when the starting point lies outside the instance contour, the radial distance is set to zero for any direction without a valid ray-contour intersection. In this paper, we take a different perspective and position polygon detection based on Polar Representation as an intermediate foreground-representation paradigm between box detection and mask segmentation.

\subsection{Representation Error}
\label{app:sec1-2}

PolarNeXt \cite{c15:polarnext} highlights an additional error in polygon detection, termed \textbf{Representation Error}. Under Polar Representation, the constructed polygons only approximate the ground-truth contours, with achievable quality strongly depending on the placement of their starting points. As a result, the uncovered regions contribute to the representation error. Moreover, the representation error is highly non-uniform and may change abruptly even under small perturbations of the starting point, making hand-crafted heuristics (e.g., center prior) unreliable for starting-point localization. In this paper, we argue that previous polar-based methods \cite{c12:polarmask,c13:polarmask++,c15:polarnext} typically select starting points on fixed feature grids based on classification scores, which inevitably limits the flexibility of Polar Representation. Therefore, one of the motivations behind our Polygon Detection Transformer is to model starting points in continuous space by leveraging contextual feature interaction.

\begin{figure*}[t]
    \centering
    \includegraphics[width=0.8\linewidth]{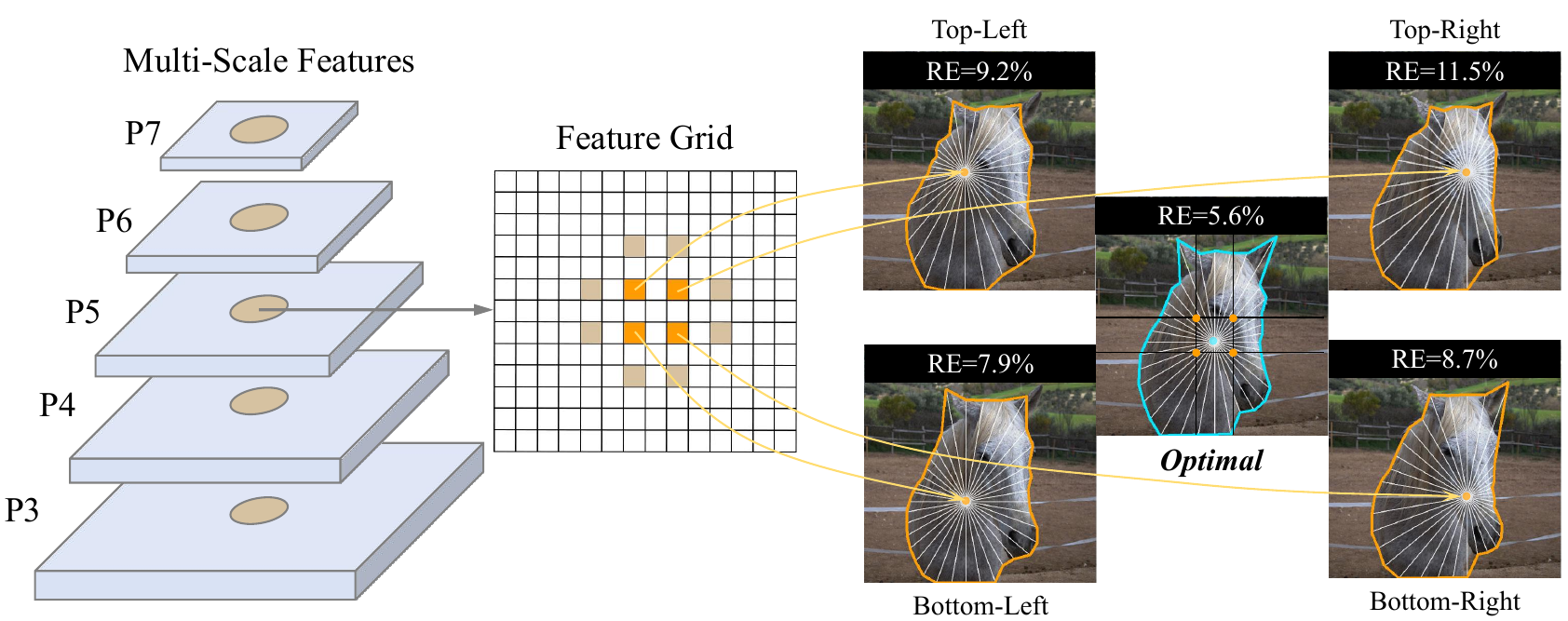}
    \caption{Motivation of continuous starting-point prediction. Four adjacent grid locations produce polygons with different representation errors (RE), while the optimal starting point lies between these discrete locations and cannot be reached by grid-based selection.}
    \label{app:fig1}
\end{figure*}

\subsection{Deformable DETR}
\label{app:sec1-3}

To reduce the computational overhead of global attention, vanilla DETR \cite{c14:detr} restricts the encoder input to low-resolution features, which is suboptimal for multi-scale detection, particularly for small objects. Deformable DETR \cite{c16:deformable-detr} addresses this issue with Deformable Attention, where a sparse set of key sampling locations is "learned" over multi-scale feature maps for feature aggregation, substantially reducing computation while accelerating convergence. This sparse-attention design has been widely adopted by modern DETR-like detectors in both encoder self-attention and decoder cross-attention. Motivated by this generality, we finally choose Deformable DETR as the baseline for Polygon Detection Transformer, whose extension naturally enables plug-and-play transfer to other modern DETR-like detectors. In addition, Deformable DETR provides a query-processing pipeline that is naturally compatible with polygon regression. To be specific, the encoder produces a set of candidate object proposals and selects high-confidence ones to initialize object queries. Then, each decoder layer predicts residual updates to the current geometric reference, which are propagated to the subsequent layer. Building on this pipeline, we instantiate Poly-DETR with two polygon-specific modules, Polar Deformable Attention and Position-Aware Training Scheme.

\section{Beyond Transformer Migration}
\label{app:sec2}

Poly-DETR is not a simple migration of Polar Representation from CNN-based detectors to Transformer architectures. As shown in Fig.~\ref{app:fig1}, the visualization illustrates the motivation for our design. Existing polar-based methods choose starting points from discrete feature-grid locations according to classification scores. In the horse example, the instance scale corresponds to the P5 feature level. The highest-scoring candidates lie on four adjacent orange grid locations, but the polygons decoded from these locations still have large representation errors. A better starting point lies between the four grid locations and therefore cannot be reached through discrete grid-based selection. Since representation error can also change sharply under very small starting-point perturbations, fixed priors and hand-crafted heuristics are unreliable for precise localization.

\begin{figure}[t]
    \centering
    \includegraphics[width=\linewidth]{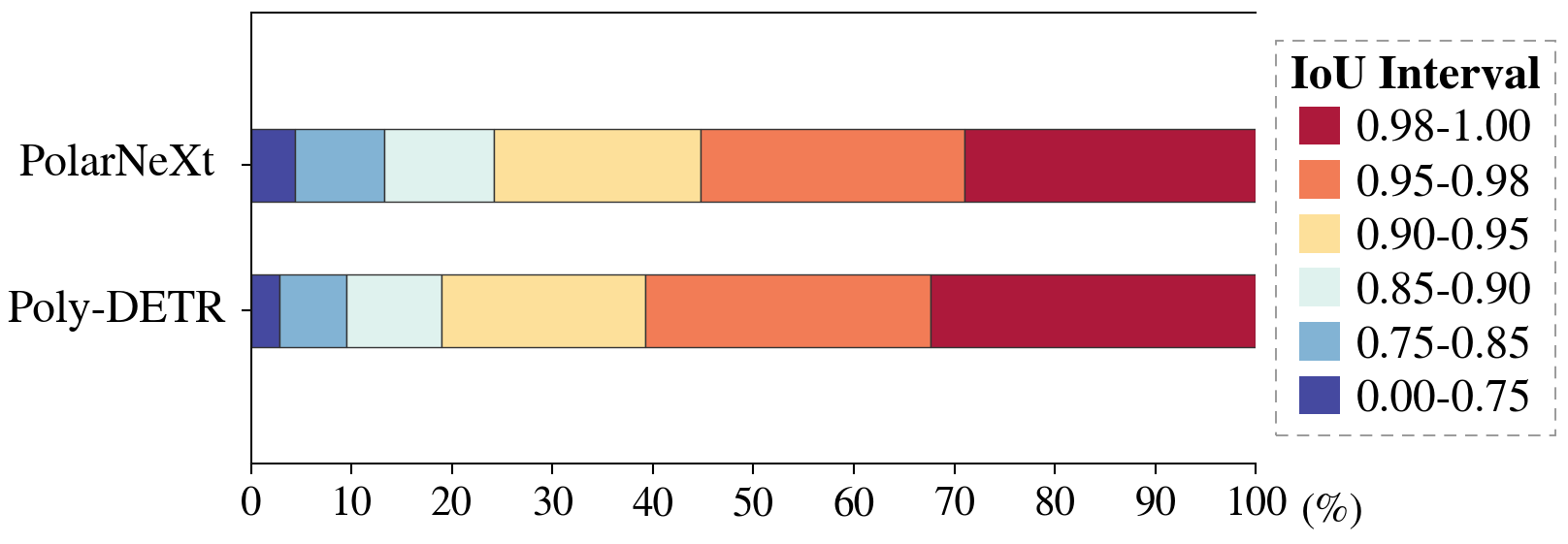}
    \caption{Distribution of representation quality for starting points predicted by PolarNeXt and Poly-DETR. Each segment represents the proportion of instances whose reference polygons fall into the corresponding IoU interval. Poly-DETR achieves a higher proportion of instances in high-IoU intervals, indicating more accurate starting-point localization.}
    \label{app:fig2}
\end{figure}

We view Detection Transformers (DETR) as a more natural remedy to this limitation, a paradigm that has become prevalent in modern object detection. Leveraging the strong global contextual feature interaction of Transformers, a sparse set of object queries can be constructed to directly regress the starting-point coordinates in continuous space, together with their corresponding radial distances. This paradigm removes the restriction to discrete grid locations, making it possible to recover starting points that yield lower representation errors. To verify this advantage, the representation-error distributions of PolarNeXt and Poly-DETR are compared in Fig.~\ref{app:fig2}. Instances are divided into intervals according to the IoU between the ground-truth contour and the polygon constructed using the predicted starting point and its reference radial distances. Poly-DETR places more instances in high-IoU intervals and fewer instances in low-IoU intervals, showing that continuous query regression reduces the representation errors caused by restricted starting-point selection. In addition, the instance-level comparisons in Fig.~\ref{app:fig5} show that Poly-DETR produces higher-quality polygons than PolarNeXt, benefiting from more favorable starting-point placement. Nevertheless, transferring Polar Representation to Detection Transformers is not straightforward and requires customized designs for polygon detection.

\section{More Details for Poly-DETR}
\label{app:sec3}

\subsection{Hybrid Supervision Strategy}
\label{app:sec3-1}

\begin{figure}[t]
    \centering
    \includegraphics[width=0.5\linewidth]{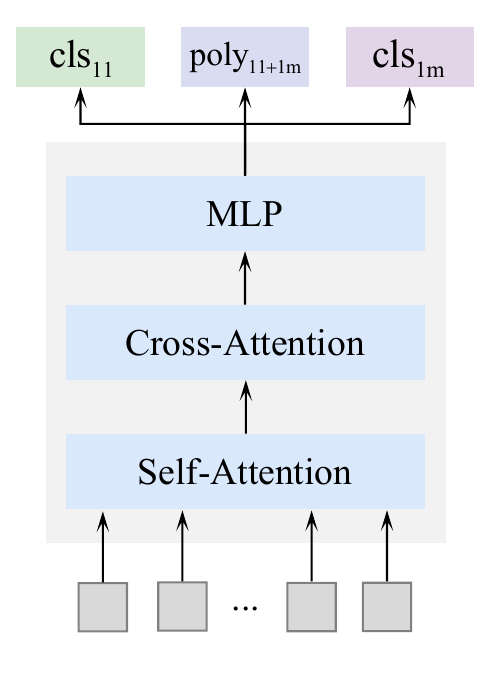}
       \caption{Architecture of Hybrid Supervision Strategy. An additional one-to-many classification branch ($\mathrm{\mathbf{cls}_{1m}}$) is introduced alongside the original one-to-one branch ($\mathrm{\mathbf{cls}_{11}}$), while the polygon regression branch ($\mathrm{\mathbf{poly}_{11+1m}}$) is shared by both supervision paths.}
    \label{app:fig3}
\end{figure}

As described in \textbf{Sec.~4.1 of the main paper}, our proposed Poly-DETR adopts Hybrid Supervision Strategy \cite{c29:msdetr} during training to accelerate convergence. Specifically, as shown in Fig.~\ref{app:fig3}, beyond the original one-to-one classification head ($\mathrm{\mathbf{cls}_{11}}$), an additional parallel one-to-many classification branch ($\mathrm{\mathbf{cls}_{1m}}$) is constructed, where multiple positive queries are assigned to each instance through one-to-many matching. Meanwhile, the polygon regression branch ($\mathrm{\mathbf{poly}_{11+1m}}$) is shared by both one-to-one and one-to-many tasks, and is supervised by the polygons associated with positive samples from both branches. In this way, the regression branch receives more effective training samples and learns the mapping from object queries to polar parameters more efficiently, enabling faster convergence under shorter training schedules. During inference, the one-to-many classification head is discarded, and only the original one-to-one branch is retained for prediction. With this design, the training schedule is reduced from 50/150 (short/long) epochs to 12/36 epochs, with only negligible performance fluctuation ($\pm$0.2 mAP). Notably, unlike \cite{c29:msdetr}, we do not swap the order of self-attention and cross-attention. 

Another point worth clarifying is that the denoising strategy in \cite{c31:dino} is also effective in Poly-DETR, but it tends to bring additional segmentation gains under longer training schedules. For a fair comparison with the baseline, we therefore adopt Hybrid Supervision Strategy as the default convergence-acceleration scheme in our main experiments.

\subsection{Inner Cost Ablation}
\label{app:sec3-2}

\begin{figure}[t]
    \centering
    \includegraphics[width=\linewidth]{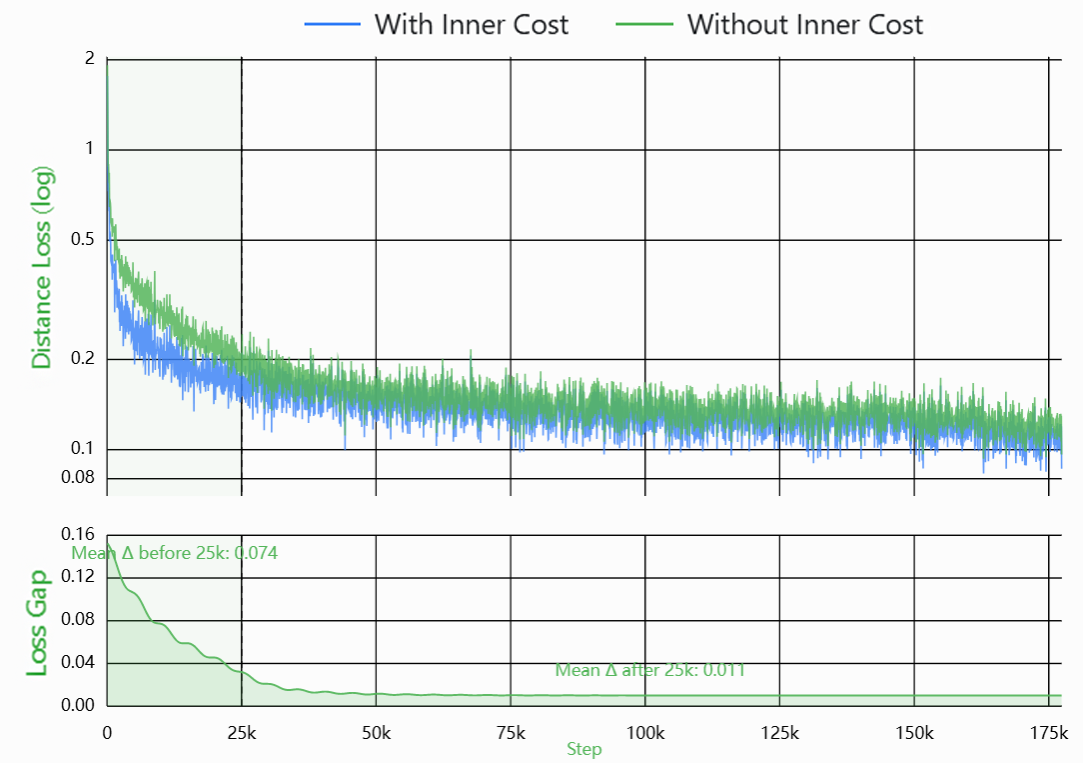}
    \caption{Training curves of Distance Loss with and without Inner Cost. The upper plot compares the Distance Loss curves, and the lower plot presents the corresponding loss gap over training iterations.}
    \label{app:fig4}
\end{figure}

As discussed in \textbf{Sec.~3.2 of the main paper}, Inner Cost is introduced to encourage object queries with starting points inside instance regions to be matched with the corresponding ground-truth contours. To further analyze its effect during training, we visualize the Distance Loss curves with and without Inner Cost in Fig.~\ref{app:fig4}. Although Inner Cost is only involved in bipartite matching and does not directly contribute gradients, it affects the selection of positive query-instance pairs, especially when the predicted starting points are still inaccurate at the early training stage. Removing Inner Cost leads to a larger Distance Loss gap during the initial optimization stage, while the difference gradually decreases as training proceeds. Nevertheless, the model trained with Inner Cost consistently maintains a slightly lower Distance Loss even after convergence. This observation suggests that Inner Cost mainly improves the quality of early-stage matching and provides more reliable dynamic supervision references for subsequent optimization.

\subsection{Effect of Query Initialization Strategy}
\label{app:sec3-3}

In polygon detection, the initialization strategy of decoder queries plays an important role in bridging encoder features and subsequent polygon regression. As shown in Tab.~\ref{app:tab1}, we investigate different geometric forms for initializing decoder object queries. When directly using polar parameters predicted from the encoder features to initialize decoder queries, the absence of Enclosing Positional Embedding (Enclosing Pos.) leads to a 0.8 mAP degradation. This indicates that preliminary polygon geometry before sufficient instance-level interaction is insufficient for stable query initialization, while the enclosing box provides a more reliable spatial prior. Alternatively, using the box-center notion to initialize parameters for object queries and positional embedding achieves comparable performance. This strategy requires retaining the original box branch as an additional prediction branch.

\begin{table}[t]
    \centering
    \begin{tabular}{ccc}
    \toprule
    \textbf{Query Initialization} &
    \textbf{Enclosing Pos.} &
    \textbf{mAP}
    \\
    \midrule
    Polar Parameters & \checkmark & 37.3\\
    Polar Parameters &             & 36.5 (-0.8)\\
    Box Parameters   &             & 37.4 (+0.1)\\
    \bottomrule
    \end{tabular}
    \caption{Effect of decoder query initialization strategies. Polar and Box parameters are predicted from the encoder features and used to initialize decoder queries. Enclosing Pos. denotes whether the enclosing rectangles of polygons are additionally provided for positional embedding.}
    \label{app:tab1}
\end{table}

\subsection{Hyperparameter Sensitivity}
\label{app:sec3-4}

\noindent \textbf{The Resolution of Rasterized Masks.} As shown in Tab.~\ref{app:tab2}, the left part studies the effect of the rasterized mask resolution. The experimental results show that using 32$\times$32 or 64$\times$64 RMask leads to nearly identical segmentation accuracy. Considering that a higher-resolution RMask is less favorable in terms of memory consumption and inference latency, we finally adopt the 32$\times$32 setting.

\noindent \textbf{The Number of Decoder Layers.} As shown in \ref{app:tab2}, the right part studies the effect of varying the numbers of decoder layers. The experimental results show that increasing the number of decoder layers fails to improve performance and instead leads to a slight degradation. By contrast, mask-based methods tend to benefit from deeper decoder stacks. We believe this is because our proposed Poly-DETR remains fundamentally a detection-style formulation, which is conceptually closer to object detection than to dense pixel-wise segmentation.

\noindent \textbf{The Number of Polar Rays.} In Poly-DETR, the number of rays in Polar Representation is set to 32, rather than the 36 rays used in other polar-based methods. This choice is driven by the design of Polar-DA, where the number of rays must follow \textbf{a power-of-two setting} so that the feature channels aggregated from all rays can be properly organized. If the number of rays deviates from this design principle, the network will fail to operate correctly.

\begin{table}[t]
\centering
\small
\setlength{\tabcolsep}{6pt}
\begin{tabular}{cccc}
    \toprule
    \multicolumn{2}{c}{RMask Resolution} &
    \multicolumn{2}{c}{Decoder Layers}\\
    \cmidrule(lr){1-2}
    \cmidrule(lr){3-4}
    \textbf{RMask} & \textbf{mAP} &
    \textbf{Layers} & \textbf{mAP}\\
    \midrule
    $32\times32$ & 37.8 &
    6 & 37.8\\
    $64\times64$ & 37.9 &
    9 & 37.7\\
    \bottomrule
\end{tabular}
\caption{
Sensitivity analysis of RMask resolution and decoder layers.
}
\label{app:tab2}
\end{table}

\section{More Details for Controlled Analysis}
\label{app:sec4}

\subsection{Mask-Based Counterpart}
\label{app:sec4-1}

To enable a systematic comparison between polygon detection and mask segmentation, we construct a mask-based counterpart \textbf{Mask-DETR} within a DETR-like framework that is kept as consistent as possible with Poly-DETR. In other words, except for the instance representation itself, most architectural components and the training recipe are matched to control confounding factors. Architecturally, the key difference is an additional mask-only branch in Mask-DETR, which is also a common ingredient in modern mask-based Transformers.

Concretely, following standard mask-based Transformer designs \cite{c5:mask2former, c6:maskdino}, Mask-DETR flattens the last three feature maps from the backbone and feeds them into the encoder to form the global memory. Meanwhile, the higher-resolution feature (at roughly $1/4$ input resolution) is retained to preserve fine-grained spatial details. This feature is fused with an upsampled version of the encoder memory to produce the mask feature, from which instance masks are predicted and then upsampled by $4\times$ to input resolution. In terms of supervision, Mask-DETR uses two mask supervisions: a pixel-wise Cross-Entropy Loss and a Dice Loss for shape consistency. Notably, these two losses play roles analogous to those in Poly-DETR: Cross-Entropy Loss is mainly responsible for enforcing interior consistency, similar in spirit to Dist L1 Loss, whereas Dice Loss serves as an exterior constraint, aligning with the role of RMask Loss. For training, Mask-DETR adopts the same data augmentation and optimization setup as Poly-DETR (i.e., random horizontal flipping and scale jittering, AdamW with a learning rate of 1e-4), and is trained with the same hybrid supervision strategy \cite{c29:msdetr} for acceleration, rather than the denoising strategy \cite{c31:dino}.

\subsection{Polygon Approximability Score}
\label{app:sec4-2}

As discussed in the main paper, polygon detection and mask segmentation have different representation characteristics. To further investigate when polygon representation becomes advantageous, we analyze the inherent polygon-friendliness of instances based on their ground-truth geometry. Specifically, we introduce a Polygon Approximability Score to quantify how well an instance can be represented by polygons under Polar Representation. This analysis motivates our further experiments on domain-specific datasets with regular-shaped instances, including PanNuke and SpaceNet.

Given an instance contour $\mathbf{C}$, a polygon $\mathbf{P}(\mathbf{s})$ can be decoded under Polar Representation from an interior starting point $\mathbf{s}=(x,y)$. Therefore, the optimal polygon is defined as:
\begin{equation}
    \mathbf{P}_{\mathrm{opt}}
    =
    \arg\max_{\mathbf{s}\in\mathcal{A}(\mathbf{C})}
    \mathrm{IoU}(\mathbf{P}(\mathbf{s}),\mathbf{C}).
\end{equation}
The corresponding Polygon Approximability Score is:
\begin{equation}
    \mathrm{AS}(\mathbf{C})
    =
    \max_{\mathbf{s}\in\mathcal{A}(\mathbf{C})}
    \mathrm{IoU}(\mathbf{P}(\mathbf{s}),\mathbf{C})
    =
    \mathrm{IoU}(\mathbf{P}_{\mathrm{opt}},\mathbf{C}).
\end{equation}
This score only depends on the ground-truth contour and measures the best achievable approximation quality under Polar Representation, assuming an optimal starting point.

To verify whether polygon approximability correlates with the effectiveness of polygon detection, we rank MS COCO instances according to their Polygon Approximability Scores and evaluate Poly-DETR and Mask-DETR on subsets with different approximability levels. As shown in Tab.~\ref{app:tab3}, the performance gap gradually increases as the subset becomes more polygon-friendly. While Poly-DETR is slightly inferior on general instances, it achieves a clear advantage on highly approximable instances. This observation suggests that polygon detection is particularly suitable for objects with compact and regular geometric structures.

\begin{table}[t]
\centering
\small
\begin{tabular}{cccc}
    \toprule
    \multirow{2}{*}{\textbf{Model}} & \multicolumn{3}{c}{\textbf{Top Subset}} \\
    \cmidrule(lr){2-4}
    & \textbf{50\%} & \textbf{30\%} & \textbf{10\%} \\
    \midrule
    Mask-DETR & 42.4 & 45.6 & 47.3 \\
    Poly-DETR & 41.8 & 46.1 & 49.2 \\
    \textit{\textbf{impr}.} & -0.6 & +0.5 & +1.9 \\
    \bottomrule
\end{tabular}
\caption{
Comparison on MS COCO subsets ranked by Polygon Approximability Score. Higher-ranked subsets contain instances that are easier to approximate by compact geometry.
}
\label{app:tab3}
\end{table}

\begin{figure*}[t]
\centering
\includegraphics[width=\linewidth]{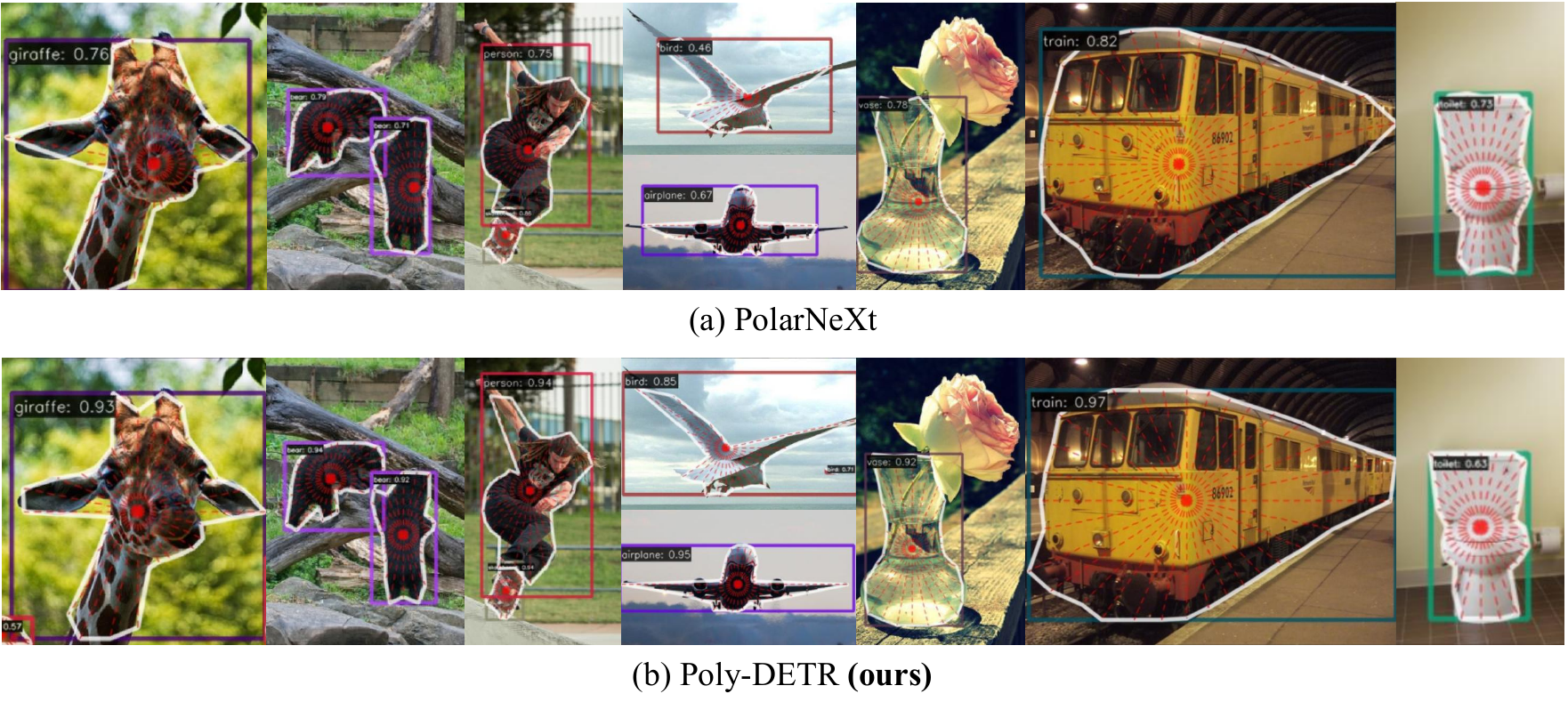}
\caption{Instance-level comparison between PolarNeXt and Poly-DETR.}
\label{app:fig5}
\end{figure*}

\section{Training Acceleration with Box Branches}
\label{app:sec5}

Poly-DETR is built upon Detection Transformers by extending the box regression of their decoders to predict polar parameters. A natural question is whether keeping the original box branch and adding a parallel polygon branch can be beneficial. The answer is yes: the box branch remains effective and can serve as a practical trick to substantially reduce training time. To make it compatible with polygon geometry, the box parameterization is slightly modified from $(x,y,w,h)$ to $(x,y,l,t,r,b)$, where $(x,y)$ denotes the coordinates of starting points instead of box centers and $(l,t,r,b)$ are distances from these starting points to their four box sides, i.e., a special case of polar parameterization with four rays.

The gain mainly comes from the matching stage. Polygon-level matching typically relies on rasterization to estimate shape overlap, which becomes expensive when computed for all Query-GT pairs (this overhead is confined to training but not required at inference). With the box branch enabled, assignment can be performed using box-based costs, largely avoiding the costly polygon matching for all pairs, while the polygon branch is still supervised by loss functions. Tab.~\ref{app:tab4} summarizes the outcome: under the same 12-epoch schedule, GPU training time is reduced from 300h to 160h, with only a minor drop in accuracy (37.8$\to$37.7 mAP). The parameter count and memory usage increase slightly (Params: 43.68M$\to$43.92M; Mems: 521MB$\to$530MB).

Finally, we emphasize that this trick is not presented as a main contribution. It introduces additional decoder complexity and re-parameterization, and practical applications typically prioritize inference accuracy, efficiency, and simplicity. We therefore include it as a practical training acceleration variant in the supplementary material, which can be useful for rapid iteration or ablation studies.

\begin{table}[t]
\centering
\small
\setlength{\tabcolsep}{4pt}
\begin{tabular}{cccccc}
    \toprule
    \textbf{Box Branch} & \textbf{mAP} & \textbf{Params} & \textbf{Mems} & \textbf{Epochs} & \textbf{GPU Time} \\
    \midrule
    \textbf{\textit{w/o.}} & 37.8 & 43.68 & 521 & 12 & 300h \\
    \textbf{\textit{w.}}   & 37.7 & 43.92 & 530 & 12 & 160h \\
    \bottomrule
\end{tabular}
\caption{Impact of retaining the box branch in the decoder.}
\label{app:tab4}
\end{table}

\section{Qualitative Visualizations}
\label{app:sec6}

\subsection{Instance-Level Comparison}
\label{app:sec6-1}

We compare the instance-level results between PolarNeXt \cite{c15:polarnext}, a state-of-the-art polar-based method, and our proposed Poly-DETR on MS COCO dataset \cite{c18:mscoco}. In the visualizations shown in Fig.~\ref{app:fig5}, we highlight the predicted starting points and the corresponding radial distances for each instance, demonstrating how the flexibility in selecting the starting point within the instance enhances Polar Representation. Poly-DETR shows superior adaptability in capturing the instance boundaries compared to PolarNeXt, especially in cases where the object geometry is complex.

\begin{figure*}[t]
    \centering
    \includegraphics[width=0.8\linewidth]{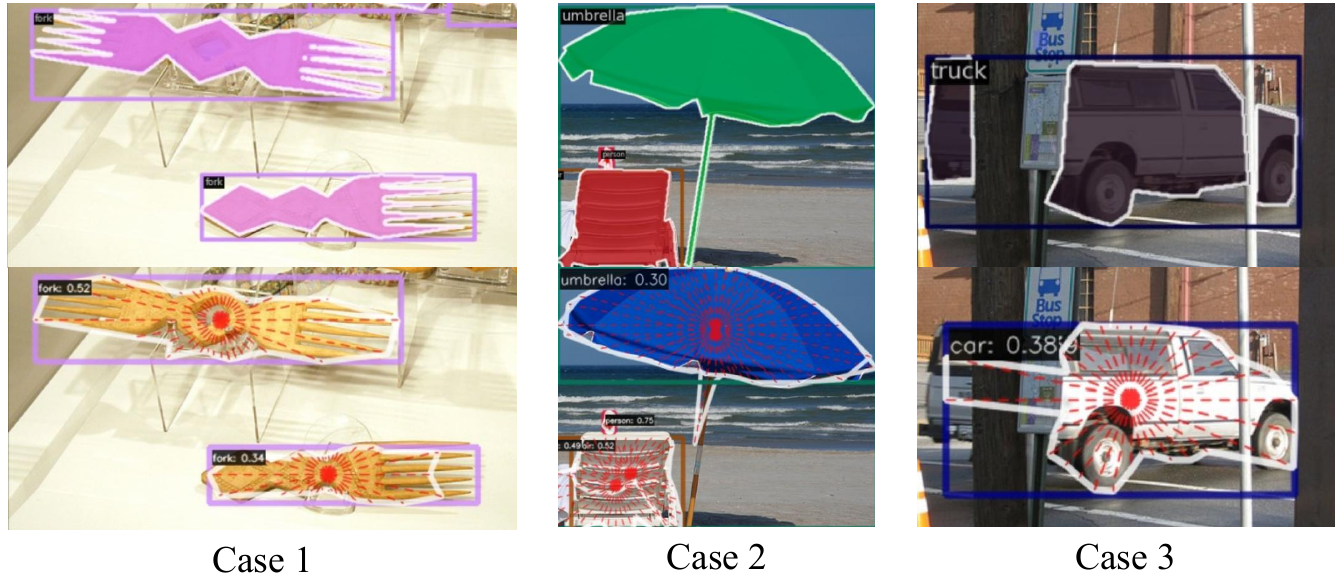}
    \caption{Failure cases of Poly-DETR.}
    \label{app:fig6}
\end{figure*}

\begin{figure*}[t]
    \centering
    \includegraphics[width=0.9\linewidth]{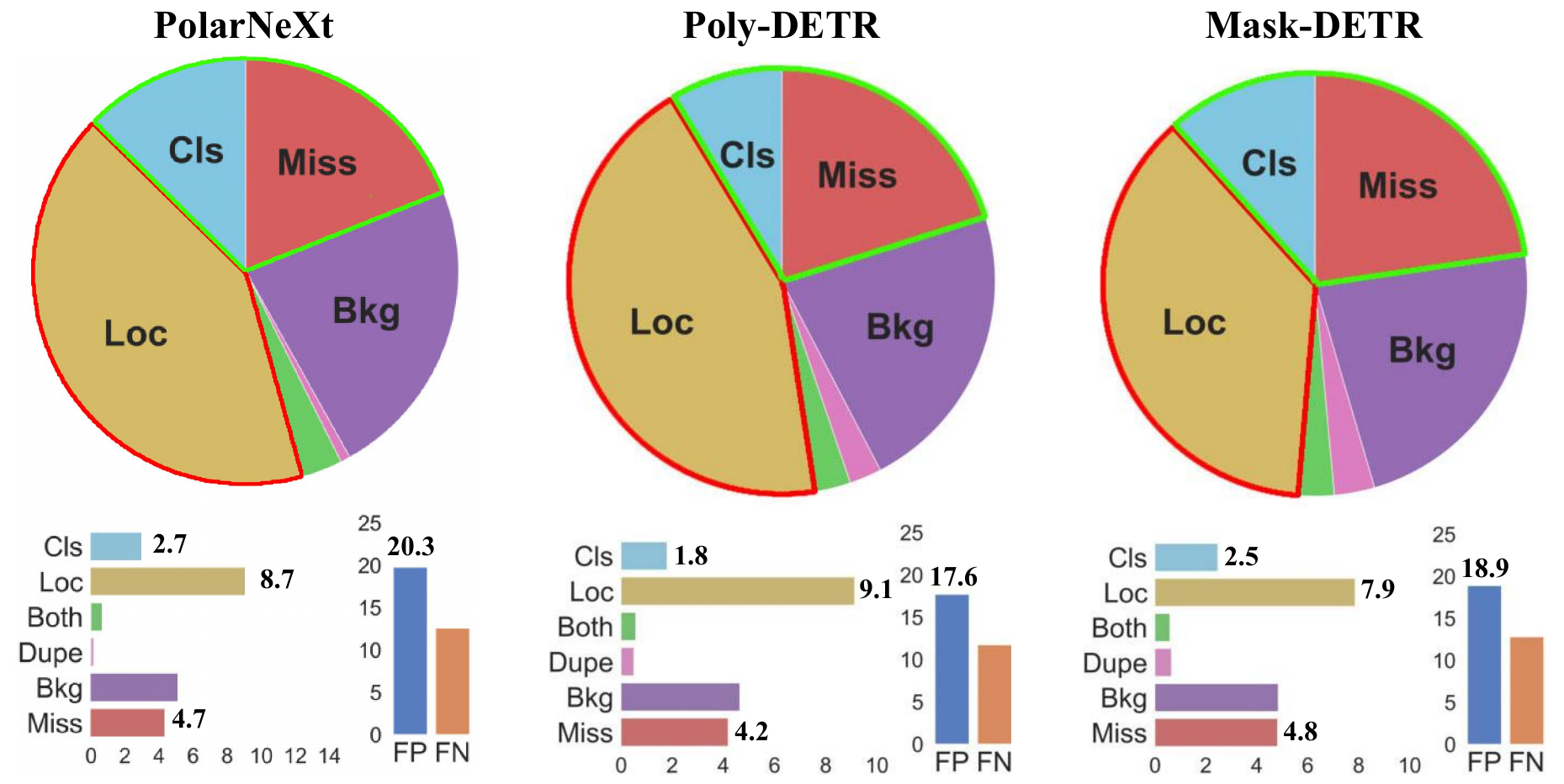}
    \caption{TIDE Error Analysis. We adopt TIDE to compare the errors of PolarNeXt, Poly-DETR, and Mask-DETR. \textit{Cls}: classification error; \textit{Loc}: localization error; \textit{Miss}: missing detections; \textit{Bkg}: background detections; \textit{Dupe}: duplicated detections; \textit{Both}: \textit{Cls}+\textit{Loc} error.}
    \label{app:fig7}
\end{figure*}

\subsection{Image-Level Results across Datasets}
\label{app:sec6-2}

\begin{figure*}[t]
\centering
\includegraphics[width=0.9\linewidth]{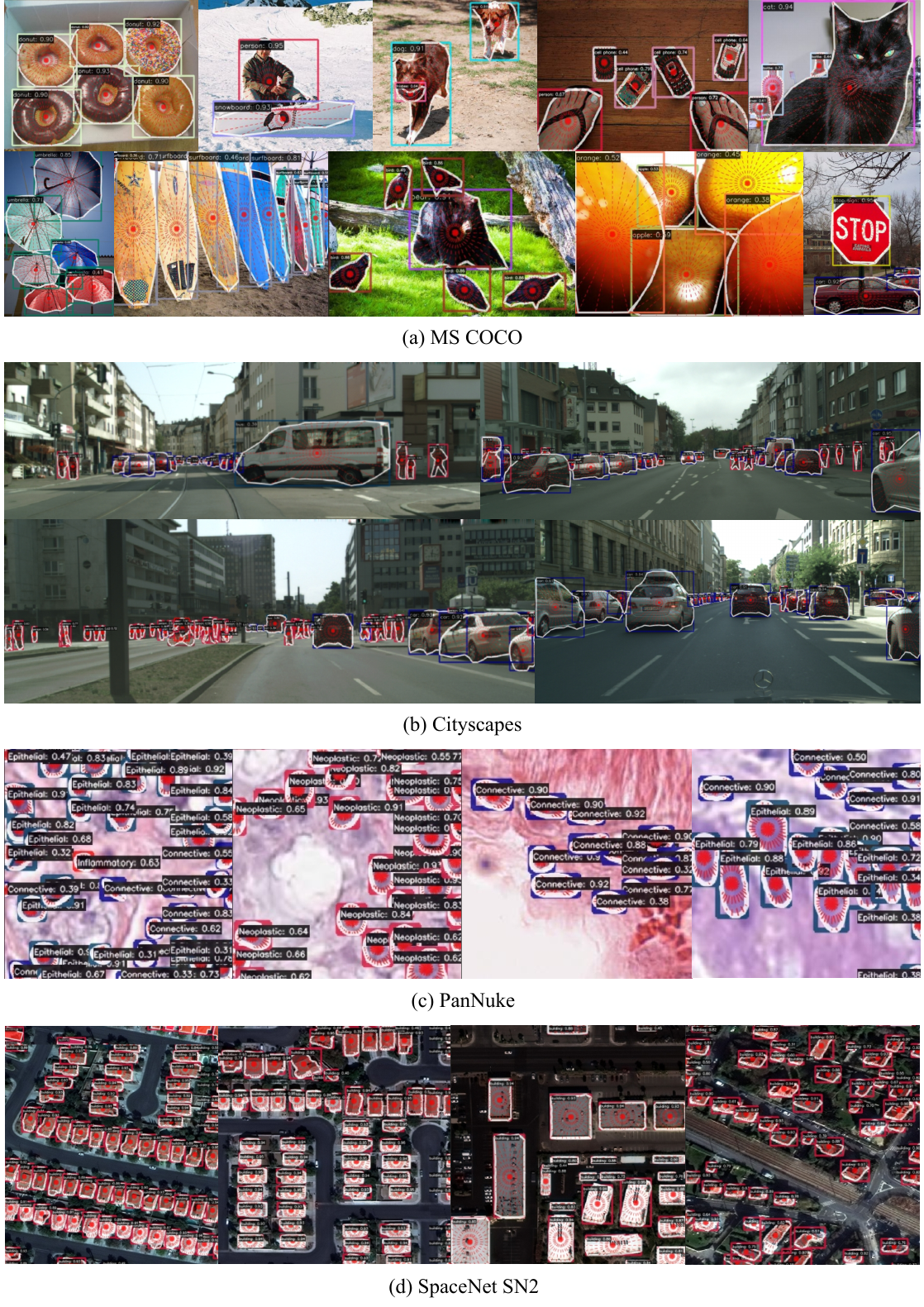}
\caption{Image-level results across four datasets.}
\label{app:fig8}
\end{figure*}

To intuitively demonstrate the effectiveness of our proposed method, we visualize the results across four datasets, as shown in Fig.~\ref{app:fig8}. The images highlight the performance of Poly-DETR on diverse datasets, including MS COCO \cite{c18:mscoco}, Cityscapes \cite{c19:cityscapes}, PanNuke \cite{c9:pannuke}, and SpaceNet SN2 \cite{c10:spacenet}. In each dataset, Poly-DETR effectively detects and represents instances, with the predicted polygons aligning closely with the instance contours.

\section{Error Analysis}
\label{app:sec7}

\subsection{TIDE Diagnosis}
\label{app:sec7-1}

We conduct a TIDE error diagnosis \cite{c43:tide} to compare PolarNeXt, Poly-DETR, and Mask-DETR, aiming to understand how different architectures and instance representations lead to distinct error profiles. As shown in Fig.~\ref{app:fig7}, PolarNeXt produces almost no duplicated detections, whereas the two Transformer-based models exhibit slightly larger Dupe errors. This difference mainly arises from their detection paradigms: PolarNeXt follows dense prediction with non-maximum suppression, which explicitly removes highly overlapping predictions, while Poly-DETR and Mask-DETR adopt NMS-free set prediction.

Under the same Transformer framework, Poly-DETR exhibits smaller classification and missing-detection errors than Mask-DETR (Cls: 1.8 vs.\ 2.5; Miss: 4.2 vs.\ 4.8), together with fewer false positives (FP: 17.6 vs.\ 18.9). We attribute this advantage to its detection-style formulation, which directly extends the original DETR prediction paradigm without introducing a separate dense mask-prediction branch. In contrast, Mask-DETR achieves a lower localization error than Poly-DETR (Loc: 7.9 vs.\ 9.1), consistent with the finer representation granularity and dense spatial supervision provided by pixel masks.

A seemingly counter-intuitive observation is that PolarNeXt also shows a slightly lower Loc error than Poly-DETR (8.7 vs.\ 9.1). However, the TIDE Loc error should not be interpreted as a direct measure of boundary quality; it reflects the potential AP gain obtained by correcting localization errors among detected instances. Compared with PolarNeXt, Poly-DETR detects more challenging instances that might otherwise be counted as missing detections. Some of these predictions are correctly classified but remain at moderate IoU and are therefore accumulated in the Loc category instead of Miss. This error redistribution is consistent with the lower Miss error of Poly-DETR and does not contradict its improved high-IoU polygon quality demonstrated in Fig.~\ref{app:fig2}. Overall, Poly-DETR reduces recognition-related errors and false alarms, while mask representation remains advantageous for fine-grained localization.

\subsection{Failure Cases}
\label{app:sec7-2}

We visualize three typical failure cases to illustrate the challenges inherent in polar-based methods. As shown in Fig.~\ref{app:fig6}, Case 1 includes two fork instances with highly variable contours, where the concave and convex areas are difficult to capture. With a limited number of rays, the network tends to overlook the concave regions in favor of maximizing the Intersection-over-Union (IoU) with the instance contour, which results in ignoring boundary details. In Case 2, the umbrella instance has both regular and protruding regions. Since the polar rays are uniformly distributed, only a limited number of rays can capture the details of the protruding region, leading to inaccuracies in Polar Representation. Case 3 highlights a car instance fragmented due to occlusion. Polar Representation typically assumes the instance as a whole around the starting point, making it difficult to handle fragmented regions and leading to incomplete predictions. We believe that future work could address these issues using adaptive polar representations and vertex refinement, among others.

\end{document}